# A Memory Optimized Data Structure for Binary Chromosomes in Genetic Algorithm

## Avijit Basak

**ABSTRACT**: This paper presents a memory-optimized metadata-based data structure for implementation of binary chromosome in Genetic Algorithm. In GA different types of genotypes are used depending on the problem domain. Among these, binary genotype is the most popular one for non-enumerated encoding owing to its representational and computational simplicity. This paper proposes a memory-optimized implementation approach of binary genotype. The approach improves the memory utilization as well as capacity of retaining alleles. Mathematical proof has been provided to establish the same.

## 1. Introduction

Genetic algorithm was introduced by John Holland as a search and optimization algorithm for unimodal and multimodal problem space. The algorithm has been used to solve numerous engineering problems of NP-complete category. The algorithm is based on few components such as encoding of the problem domain as chromosomes, a fitness function for selection of individual chromosome, crossover for exchange of allele values between chromosomes and mutation to alter randomly selected alleles. Being superior to gradient descent, genetic algorithm has been used for optimization in unimodal and multimodal problem space along with training of machine learning models like neural network. However, GA also suffers from the drawback of excessive resource consumption owing to its population-based approach for optimization.

For several decades the algorithm has been improved by numerous researchers in different ways. Improvements have been done for selection, crossover and mutation operators. The adaptive GA was also introduced for the same purpose. But there is not much work related to minimization of resource consumption which has been a bottleneck towards its use in large scale problems. This work is an endeavour to minimize memory consumption for binary encoding.

## 2. Utility and Implementation of Binary Encoding
### 2.1. Schema Theorem and Binary Encoding

Holland's Schema theorem provides the mathematical foundation for evolutionary dynamics of Genetic algorithm. In Schema theorem, a *schema* is defined as a template that identifies a subset of chromosomes with similarities at certain allele positions. Acceptable allele values in a schema are the set of alphabets used in the genotype and the wildcard character * which represents don't care symbol. So, for a genotype of length L and alphabet cardinality k there can be maximum $(k + 1)^L$ schemata as described in Equation 1. Schema theorem states, **schemas** having *lower defining length*, *lower order* and *above average fitness* increases exponentially in frequency over generations. The *order* of a *schema* is defined as the number of fixed positions in the template and the distance between first and last specific alleles of a schema is termed as *defining length*. The theorem is expressed as an equation (Equation 2) given below:

**Equation 1:**

Maximum possible schemata count = $(k + 1)^L$

where, k = cardinality of genotype

L = Length of chromosome

**Equation 2:**

$$E(m(H, t + 1)) = \frac{m(H,t) * favg(H)}{favg}(1 - \frac{\delta(H)}{L-1}p_c)(1 - p_m)^{o(H)}$$

where, m(H, t) = Number of chromosomes which belong to schema H, at generation t

favg(H) = Observed average fitness of the schema H

favg = Observed average fitness of the population at generation t

L = Length of chromosome

$p_c$ = Probability of Crossover

δ(H) = Defining length of schema

o(H) = Order of Schema

$p_m$ = Probability of mutation of each allele

In the above equation, the ratio $\frac{\delta(H)}{L-1}p_c$ determines the disruption probability of a schema due to crossover and $p_m$ determines the probability of disruption due to mutation of an allele. The ratio $\frac{\delta(H)}{L-1}p_c$ implies *for a given defining length $\delta(H)$, higher length of chromosome ensures lower minimum probability of schema disruption,* which eventually improves the quality of optimization. However, the quality also depends on the number of schemata processed over generations during convergence. *Higher number of processed schemata ensures better quality of search result*. According to Equation 1 higher chromosome length also ensures more possible schemata count.

In any real-life problem choice of genotype decides the alphabet cardinality and hence length of the chromosome. Lower cardinality ensures higher length of chromosome. Due to minimum cardinality, use of binary encoding thus ensures processing of maximum possible schemata over generations with minimum probability of schema disruption due to crossover. This makes binary genotype an obvious choice for any critical search and optimization problem.

## 2.2. IMPLEMENTATION OPTIONS AND MEMORY UTILIZATION

There are several approaches for implementation of a binary chromosome. Few obvious approaches are to represent it as an array of *Character*, *Boolean* or *Integral* data type and store each binary allele as one array element. All these approaches consume minimum 1 byte of memory for a single binary allele. This results in memory utilization of **12.5**% eventually wasting **87.5**% of memory. The amount of unused memory increases with problem dimension. This becomes a bottleneck for problem domain with millions of dimensions and makes the usage of binary encoding nearly impossible.

## 3. METADATA-BASED IMPLEMENTATION & BENEFITS
### 3.1. IMPLEMENTATION APPROACH

The *metadata-based* implementation of binary chromosome proposed here minimizes the problems mentioned in the previous section. A binary chromosome is represented as an array of an integral data type. The *first element* of the array should contain the *number of alleles* used in this chromosome and represent the *metadata*. Rest of the elements in the array should contain binary allele information as bits. Each single bit represents a binary allele. The data type can be either signed or unsigned. However, MSB of each element should not be used to retain allele data to avoid any semantic conflict of the underlying implementation. The number of array elements (m) required to represent a chromosome of length L is expressed by Equation 3 given below.

**Equation 3:**

$$m = 1 + roundup\left(\frac{L}{n}\right)$$

where, L = Length of chromosome

n = Number of unsigned bits in data type

A sample chromosome containing randomly generated binary values of length 10 is shown below in a tabular form in Figure 1. Unsigned Byte is used as the elementary data type of the array. The first row represents the indexes of array elements. The second row represents the index of each binary allele within the element and the bottom row contains actual data. The first element of the array retains the length of the chromosome as metadata and the rest of the elements contain the actual allele information as placed within the respective cells. Few of the bits (marked as X) of the second array element might be left unused depending on the length of the chromosome. Similar representation using the signed Byte is shown in Figure 2. The difference between these two representations can be noted for MSB of the 3rd array element which is left unused for signed data type.

| Array Element Index | 0 | | | | | | | | 1 | | | | | | | | 2 | | | | | | | |
|---|---|---|---|---|---|---|---|---|---|---|---|---|---|---|---|---|---|---|---|---|---|---|---|---|
| Bit Index in Array Element | 0 | 1 | 2 | 3 | 4 | 5 | 6 | 7 | 0 | 1 | 2 | 3 | 4 | 5 | 6 | 7 | 0 | 1 | 2 | 3 | 4 | 5 | 6 | 7 |
| Allele Information | | | | | 1 | 0 | 1 | 0 | X | X | X | X | X | X | 1 | 0 | 1 | 0 | 0 | 1 | 1 | 0 | 0 | 0 |

**Figure 1**

| Array Element Index | 0 | | | | | | | | 1 | | | | | | | | 2 | | | | | | | |
|---|---|---|---|---|---|---|---|---|---|---|---|---|---|---|---|---|---|---|---|---|---|---|---|---|
| Bit Index in Array Element | 0 | 1 | 2 | 3 | 4 | 5 | 6 | 7 | 0 | 1 | 2 | 3 | 4 | 5 | 6 | 7 | 0 | 1 | 2 | 3 | 4 | 5 | 6 | 7 |
| Allele Information | | | | | 1 | 0 | 1 | 0 | X | X | X | X | X | 1 | 0 | 1 | X | 0 | 0 | 1 | 1 | 0 | 0 | 0 |

Figure 2

## 3.2. PSEUDOCODE

Pseudocode for the proposed approach has been provided below.

```
//Pseudocode of Genetic Algorithm for Onemax domain
//An abstract integral data type 'D' of size n bits is used here
procedure GENETIC-ALGORITHM(N, pc, pm, L)
    parameter(s):
        N - population size
        pc - probability of crossover
        pm - probability of mutation
        L - length of chromosome
    output: best individual of converged population

    //Initialization:
    t ← 0
    P(t) ← INITIALIZE(N, L, n)
    //procedure EVALUATE-FITNESS evaluates each chromosome's fitness
    EVALUATE-FITNESS(P(t))

    //Execute Optimization:
    while (termination condition not met)
        count = 0
        Ptmp = P(t)
        while(count < N/2)
            //procedure 'SELECT-CHROMOSOME-PAIR' selects
            //a pair of chromosomes using specific selection method
            //like Roulette Wheel or Tournament selection
            C1, C2 = SELECT-CHROMOSOME-PAIR(Ptmp)
            PERFORM-CROSSOVER(C1, C2, pc, L, n)
            MUTATE(C1, pm, L, n)
            MUTATE(C2, pm, L, n)
            count++
        END while

        P(t+1) ← Ptmp
        t ← t + 1
        EVALUATE-FITNESS(P(t))
    END while

    return (best individual from P(t))
END procedure
```

```
//This procedure initializes the population of chromosomes randomly
procedure INITIALIZE (N, L, n)
    parameter(s):
        N - population size
        L - length of chromosome
        n - no of unsigned bit in data type
    output: initialized population

    dim = CALCULATE-ARRAY-DIM(L, n)
    Initialize an empty population P

    count = 0;
    do
        //initialize an array of abstract integral data type 'D'
        D[] chromosome = new D[dim + 1]

        //Operation 'generate-mask' generates mask of length n.
        //Mask should have leading '0' of given offset count
        //and rest of the elements should be '1'.
        offset = n - (L % n)
        //mask for the first element
        mask_first = generate-mask(offset, n)
        //mask for the rest of the elements
        mask_other = generate-mask(0, n)

        chromosome[0] = L //initialize length of chromosome as metadata
        //initialize the first element
        chromosome[1] = mask_first AND random(Maximum(D))
        for(i = 1; i < dim; i++)
            //initialize rest of the elements
            chromosome[i + 1] = mask_other AND random(Maximum(D))

        Add the chromosome to existing population P
        count++

    while(count < N)

    return P
END procedure

procedure CALCULATE-ARRAY-DIM(L, n)
    parameter(s):
        L - length of chromosome
        n - bit size of datatype
    output: array dimension

    dim = 1 + roundup(L / unsigned-bit-count(n))

    return dim
END procedure
```

```
//This procedure performs a one point crossover
procedure PERFORM-CROSSOVER (C1, C2, pc, L, n)
    parameter(s):
        C1 - chromosome 1
        C2 - chromosome 2
        pc - probability of crossover
        L - length of chromosome
        n - no of unsigned bit in data type
    output - pair of chromosomes after crossover

    offset = n - (L % n)
    crossover_index = random(L) + offset
    chromosome_array_index = round-down(crossover_index/n) + 1
    allele_index = crossover_index % n

    //operation 'exchange' exchanges bits between two chromosomes
    // C1 and C2 from begining to allele present at 'allele_index'
    //of array element present at index 'chromosome_array_index'
    exchange(C1, C2, chromosome_array_index, allele_index)

    return (C1, C2)
END procedure

//This procedure mutates a chromosome with certain probability
procedure MUTATE (C, pm, L, n)
    parameter(s):
        C - chromosome
        pm - probability of mutation
        L - length of chromosome
        n - no of unsigned bit in data type
    output - chromosomes after mutation

    offset = n - (L % n)
    index = 0;
    while(index < L)
        if random(1) < pm
            mutation_index = offset + index
            chromosome_array_index = round-down(mutation_index/n) + 1
            allele_index = mutation_index % n

            //operation 'flip' inverts a bit of chromosome 'C'
            // located at mentioned 'allele_index' of the
            //array element at index 'chromosome_array_index'.
            flip(C, chromosome_array_index, allele_index)
        END if

        index++
    END while

    return C
END procedure
```

### 3.3. BENEFITS

The proposed implementation approach is effective both in accommodating more alleles in a chromosome as well as ensuring better memory utilization. In a conventional representation of binary chromosome as character or Boolean data type, the number of alleles in a chromosome is limited to the maximum number of elements an array can hold. Contrary to that in this newly proposed approach number of alleles is dependent on the maximum number of bits an array can contain depending on the data type used as well as the maximum positive number the corresponding data type can represent, whichever is smaller. Memory utilization also depends on chromosome length and data type as represented by the equations (Equation 4 & 5) given below. The tables (Table 1 & 2) list calculated values of the same for different data types (signed and unsigned) of varying capacity which clearly demonstrates the superiority of the proposed approach. The maximum number of elements in an array is considered as $2^{16}$ for the comparison. The results clearly demonstrate that the proposed scheme improves the memory utilization to a considerable extent and higher capacity of data type ensures better utilization of memory.

**Equation 4:**

Maximum Possible Chromosome Length = $\min(2^n - 1, n*(M - 1))$

where, n = No of unsigned bits in data type

M = Maximum element capacity of an array

**Equation 5:**

Memory Utilization = $\dfrac{L}{m*N}$

where, L = length of chromosome

m = No of array elements required for the chromosome

N = No of bits in data type

**Table 1**

| UNSIGNED DATATYPE CAPACITY (bit) | MAX POSITIVE NUMBER in DATATYPE (METADATA) | ARRAY CAPACITY (bit) | MAX CHROMOSOME LENGTH | MEMORY UTILIZATION for MAX CHROMOSOME LENGTH (Percentage) |
|---|---|---|---|---|
| 8 | 255 | 524280 | 255 | 96.59% |
| 16 | 65535 | 1048560 | 65535 | 99.97% |
| 32 | 4294967295 | 2097152 | 2097152 | 99.99% |
| 64 | 18446744073709551615 | 4194304 | 4194304 | 99.99% |

Table 2

| SIGNED DATATYPE CAPACITY (bit) | MAX POSITIVE NUMBER in DATATYPE (METADATA) | ARRAY CAPACITY (bit) | MAX CHROMOSOME LENGTH | MEMORY UTILIZATION for MAX CHROMOSOME LENGTH (Percentage) |
|---|---|---|---|---|
| 8 | 127 | 458752 | 127 | 79.37% |
| 16 | 32767 | 983040 | 32767 | 93.68% |
| 32 | 4294967295 | 2031616 | 2031616 | 96.87% |
| 64 | 18446744073709551615 | 4128768 | 4128768 | 98.43% |

## 4. CONCLUSION

Genetic Algorithm is proven to be very effective compared to gradient descent for finding global optimum in both unimodal and multimodal problem space. Considering the diverse domain of machine learning GA has immense potential of being utilized both in classical machine learning as well as deep learning. Due to its population-based approach, excessive memory consumption of GA is considered as a bottleneck towards its use in large scale problem. The implementation approach presented here would minimize memory wastage. Better memory utilization has been proven mathematically, which is improved from **12.5%** to around **90%**. Pseudocode is also provided to demonstrate the feasibility of the approach.

## REFERENCES:

[1] DeJong, K. A., An analysis of the behaviour of a class of genetic adaptive systems Ph.D. dissertation, University of Michigan (1975)

[2] Goldberg, D. E., Genetic Algorithms in Search, Optimization and Machine Learning, Addition-Wesley Publishing Co., Inc. Boston, MA, USA. (1989)

[3] Mille, Brad L., Goldberg, D. E., Genetic Algorithms, Tournament Selection, and the Effects of Noise, Complex Systems 9 (1995) 193- 212

[4] Srinivas, M. and Patnaik, L. M., Adaptive Probabilities of Crossover and Mutation in Genetic Algorithms, IEEE Transactions on Systems, Man and Cybernetics, VOL. 24, NO. 4, APRIL 1994.

[5] Goldberg, D.E., Deb, K., A Comparative Analysis of Selection Schemes Used in Genetic Algorithms, University of Illinois at Urbana-Champaign, Urbana, United States

[6] Kühn, M., Severin, T., Salzwedel H., Variable Mutation Rate at Genetic Algorithms: Introduction of Chromosome Fitness in Connection with Multi-Chromosome Representation, International Journal of Computer Applications (0975 – 8887), Volume 72– No.17, June 2013

[7] Korejo++, I.A., Khuhro, Z.U.A., Jokhio, F. A., Channa*, N., Nizamani, H. A., An Adaptive Crossover Operator for Genetic Algorithms to Solve the Optimization Problems, Sindh University Research Journal (Science Series) Vol.45 (2) 333-340 (2013) Ding, W. and Marchionini, G. 1997 A Study on Video Browsing Strategies. Technical Report. University of Maryland at College Park.